# High Accuracy Classification of White Blood Cells using TSLDA Classifier and Covariance Features

Hamed Talebi[1]*, Amin Ranjbar[2], Alireza Davoudi[1], Hamed Gholami[1], Mohammad Bagher Menhaj[2]

**Abstract— creating automated processes in different areas of medical science with the application of engineering tools is a highly growing field over recent decades. In this context, many medical image processing and analyzing researchers use worthwhile methods in artificial intelligence, which can reduce necessary human power while increases accuracy of results. Among various medical images, blood microscopic images play a vital role in heart failure diagnosis, e.g., blood cancers. The prominent component in blood cancer diagnosis is white blood cells (WBCs) which due to its general characteristics in microscopic images sometimes make difficulties in recognition and classification tasks such as non-uniform colors/illuminances, different shapes, sizes, and textures. Moreover, overlapped WBCs in bone marrow images and neighboring to red blood cells are identified as reasons for errors in the classification task. In this paper, we have endeavored to segment various parts in medical images via Naïve Bayes clustering method and in next stage via TSLDA classifier, which is supplied by features acquired from covariance descriptor results in the accuracy of 98.02%. It seems that this result is delightful in WBCs recognition.**

## 1. Introduction

Segmentation of white blood cells is defined as partitioning cells from its neighboring volume. WBCs segmentation in microscopic images is a complicated procedure and always makes difficulties in attaining a standard technique to yield robust results. Accordingly, some conventional methods are reviewed in this area:

### 1.1.Threshold-based methods

These methods are intensively dependent on the uniformity of images in the database. In this case, reaching to high-speed processing with acceptable results is feasible [6,18], but in the presence of artifacts and unwanted noises, it results in exceeded segmentation and seems to be inefficient. There have been several techniques to threshold any image, but this always considered as the first processing stage in most microscopic image analyses [3].

In [7], the Otsu thresholding method has been applied to segment nuclei of WBCs and then with the approximated location of the nucleus and active contour method, an estimated border around cytoplasm is extracted. In [8], by Otsu thresholding that works in circular histogram iteratively, have been tried to segment WBCs. In this paper, the Otsu method is applied to the H and S components of the medical images transformed into the HSI color space.

Huang et al. in [2], to improve image analysis, used both G channel from RGB color space and saturation channel from the HSV model. Due to the presence of darker WBCs in the G channel and more brilliant cells in the saturation channel, it helps to improve the intensity of WBCs by making a ratio of saturation and green components. In this case, due to less number of pixels related to nuclei in the histogram of the image, the two-class Otsu method is not succeeded to discriminate desired segments. Moreover, Gautam and Bhadauria in [9], after preprocessing tasks, including contrast improvement and histogram equalization, they also used Otsu thresholding to extract WBCs in microscopic blood image. In [10], Prinyakupt et al. proposed a method based on averaging blue and red components and then dividing into its green channel components. In this regard, they tried to draw different axial lines from the center point in each nucleus, which were applied in the method proposed in [11]. This generates an oval around the desired cytoplasm. This method could achieve 92.9% and 94.7% for discrimination of nuclei and cytoplasm, respectively.

[1] Department of Software Engineering, Amirkabir University of Technology, Tehran, Iran
[2] Department of Electrical Engineering, Amirkabir University of Technology, Tehran, Iran
* Corresponding author's email: hamed.talebi.aut@gmail.com

In [13], Zheng et al. proposed a method based on the watershed algorithm that causes over-segmentation. To solve this, they applied a threshold condition, which merges two neighbors are closely similar to each other. This method results in the accuracies of 97.51% and 91.04% for nuclei and cytoplasm respectively.

### 1.2. Pattern recognition-based methods

In all methods use pattern recognition, the WBCs are figured as desired objects [13]. These methods are categorized into two general groups, including supervisory classification and unsupervised clustering. The common methods in this topic are included k-means [5,14,17], fuzzy c-means [15,19,20,24] and Expectation-Maximization (EM) [16] clustering methods.

In this regard, wide verity of the methods in cell segmentation profit from fuzzy inference systems. In [21], Ghosh et al. proposed an approach based on fuzzy divergence to test Gamma, Gaussian, and Cauchy membership functions to separate blood cells. They also in another paper [22,23], by calculation of fuzzy divergence based on Renyi entropy could reach an accuracy of 96.18%.

For unsupervised methods, Pan et al. in [4] proposed a technique using a simulated visual attention system to extract WBCs in blood microscopic images. They used morphological operators and support vector machines ensemble to generate samples and then learning efficiently. In another paper from Pan et al., they used an Extreme Learning Machine (ELM) instead of SVM classifier, which resulted in a higher accuracy [12].

### 1.3. Active contour-based methods

These methods are grouped into parametric models and geometric models. Generally, active contour-based methods suffer from overlapped objects in the image. Additionally, the presence of any unwanted object in the picture makes analyses more complicated and decline prediction performance [3].

Co et al. in [25] proposed a technique to combine nuclei via the mean-shift method. They applied a Gradient Vector Flow (GVF) technique to distinguish and to remove undesirable borders, but unfortunately, the algorithm needs high processing time and could reach to 88.35% and 67.66% to segment nuclei and cytoplasm correspondingly. In 2009, Tofighi et al tried to extract WBCs via Gram Schmidt Orthogonality method [1,26]. In this method, the pixels are close to desired objects illustrated more brilliant and contrary unwanted pixels with the low-intensity pixel. Subsequently, the Otsu thresholding is applied on image histogram to classify WBCs. In another work from Tofighi et al. in 2011, they improved their algorithm via active contours based snake method to define boundaries of cytoplasm in WBCs [27].

In this paper, we have tried to develop a technique to detect different types of WBCs, as depicted in Fig. 1. It seems the proposed method can achieve remarkable results in WBCs detection in comparison to other previous methods.

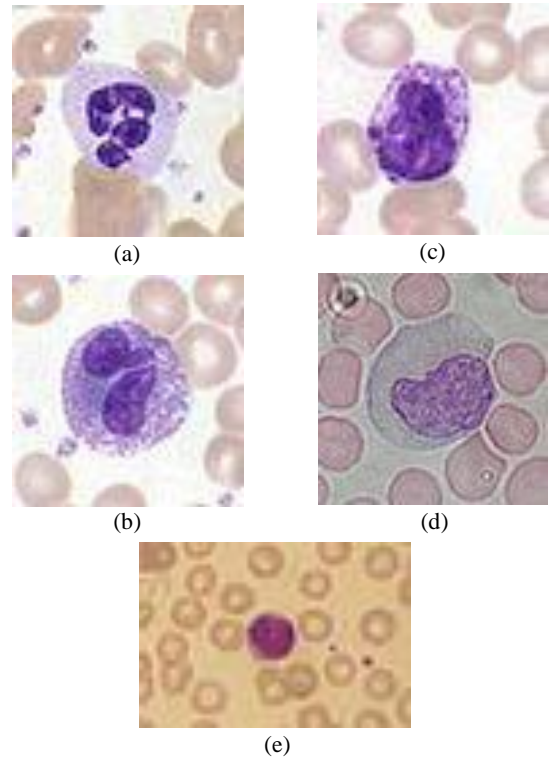

**Fig. 1.** Different types of white blood cells: (a) Neutrophil, (b) Eosinophil, (c) Basophil, (d) Monocyte, (e) Lymphocyte

## 2. Materials and methods

### 2.1. Segmentation

Our method proposes different tasks in preprocessing and processing phases to achieve high performance in WBCs recognition. Different tasks in our algorithm are described as following:

*1.* Firstly we try to convert our image database to "Lab" color space. This space can represent alterations of different colors in all images efficiently. Additionally, the next advantage is the "L" channel is fully independent of "a" and "b" components. Thus, for this conversion, the following equations can be used:

$$\begin{bmatrix} X \\ Y \\ Z \end{bmatrix} = \frac{1}{0.17697} \begin{bmatrix} 0.49 & 0.31 & 0.2 \\ 0.17697 & 0.8124 & 0.01063 \\ 0 & 0.01 & 0.99 \end{bmatrix} \begin{bmatrix} R \\ G \\ B \end{bmatrix} \qquad (1)$$

We use the equations (2), (3), and (4) to convert color coordination from XYZ to Lab:

$$L = 116 f\left(\frac{Y}{Y_0}\right) - 16 \qquad (2)$$

$$a = 500\left(f\left(\frac{X}{X_0}\right) - f\left(\frac{Y}{Y_0}\right)\right) \qquad (3)$$

$$b = 200 \left( f\left(\frac{Y}{Y_0}\right) - f\left(\frac{Z}{Z_0}\right) \right) \qquad (4)$$

The equations of (2), (3) and (4) get initialized as $Y_0 = 100, X_0 = 95.047, Z_0 = 108.883$ which are called reference white color coordination. The mapping function $f(t)$ is also defined according to equation (5):

$$f(t) = \begin{cases} t^{\frac{1}{3}} & if\ t > \left(\frac{6}{29}\right)^3 \\ \frac{1}{3}\left(\frac{29}{6}\right)^2 t + \frac{4}{29} & o.w. \end{cases} \qquad (5)$$

*2.* Secondly, due to more information of WBCs in channel "a" of the color space, it was quite reasonable to continue the proposed algorithm through this channel.

*3.* At the third stage in images preprocessing phase, contrast improvement is a mandatory task, which makes images more informative and brings them into the same situation to analyze.

Thus, to attain this, the histogram equalization, which is the most popular technique among image enhancement methods comes into account.

*4.* At the current stage, we try to segment images into different regions. Hence, the Naïve Bayes estimator is adjusted to split images into three distinct sections.

*5.* Morphological operations are applied to remove unwelcome particles and noises which lead to an appropriate model for labeling.

In Fig. 2, the results of the algorithm to segment the input image is illustrated.

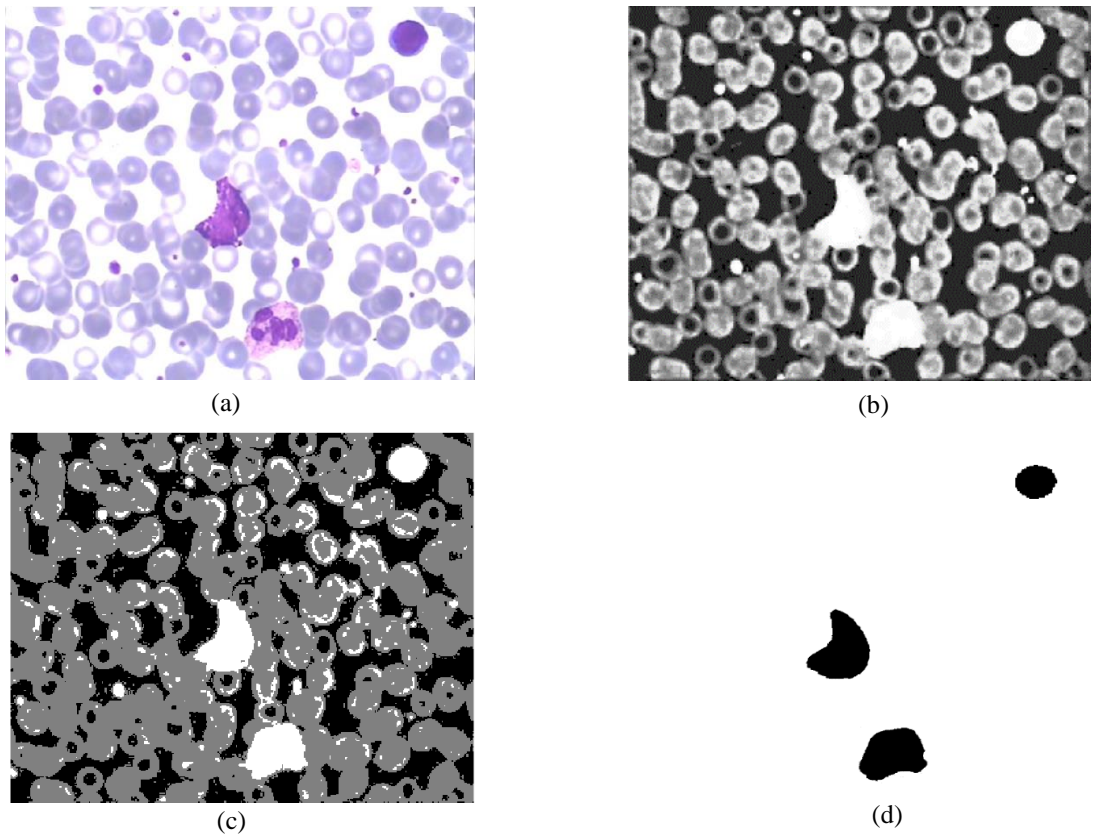

(a)

(b)

(c)

(d)

**Fig. 2.** Proposed segmentation procedure: (a) Raw microscopic input image, (b) "L" channel of the image through Lab conversion, (c) After applying Naïve Bayes clustering, (d) Applying some morphological operations

## 2.2. Feature Extraction

### 2.2.1. SIFT features

SIFT features are one of the significant findings by researchers in machine vision over decades. The main reasons for the popularity of SIFT are:

- Generation of a dense descriptor with the length of 64 or 128 pixels
- Rapid implementation

- More robust than other transformations.

These features are applied to find the location of desired objects in any sequence of images.

### 2.2.2. PHOW features

PHOW features are categorized into D-SIFT descriptors [1], which are evaluated in different scales. Meanwhile, this method produces too many features which should use the Bag of Words (BoW) technique to reduce the number of calculated features. PHOW increases the speed of feature extraction, while the feature matrix is the same.

### 2.2.3. Covariance descriptor

Here we attempt to summarize covariance descriptor theory and its application mainly in WBCs recognition. In this content, the symbol "I" represents the input image, which can be intensity or colored image. Additionally, "F" stands for feature image extracted from the input image with the dimension of $W \times H \times d$ and is defined as equation (6).

$$F(x, y) = \varphi(I, x, y) \tag{6}$$

The function of $\varphi$ can be any function like gradient, filter responses, etc. For the rectangular region of $\{Z_i\}$ where $i = 1, \dots, S$, the d-dimensional feature points are placed into region R. In other words, R shows covariance matrix with the size of $d \times d$ from feature points.

$$CR = \frac{1}{S-1} \Sigma_{i=1}^{S}(Z_i - \mu)(Z_i - \mu)^T \tag{7}$$

In equation (7), $\mu$ stands for mean value on all feature points. To solve the WBCs classification problem, the function $\varphi(I, x, y)$ is calculated as the following:

$$\varphi = \left[ x \, y \, |I_x| \, |I_y| \sqrt{I_x^2 + I_y^2} \, |I_{xx}| \, |I_{xx}||I_{yy}| \, \arctan \frac{|I_x|}{|I_y|} \right]^T \tag{8}$$

In the previous equation, x and y are pixel location and $I_x, I_{xx}, \dots$ represent derivatives of the intensity image, and the last term shows the direction of edges in 2D space. By this mapping, the input image is transformed into n-dimensional space.

Hence, the covariance descriptor is defined as an $8 \times 8$ matrix, which due to symmetric properties; only the upper section is kept for next manipulations.

## 3. Riemannian Theory

### 3.1. Overview

The space of all symmetric matrixes with the size of $n \times n$ and located in the space of real square matrix $M(n)$ is defined as $S(n) = \{S \in M(n), S^T = S\}$. The matrixes are symmetric positive definite if:

$$P(n) = \{P \in S(n), u^T Pu > 0, \forall u \in \mathbb{R}^n\}$$

These matrixes are capable of being diagonalized while its Eigenvalues are positive definite.

At the first step, to calculate exponential and logarithmic matrixes of P, the decomposition is evaluated over P:

$$P = U diag(\sigma_1, \dots, \sigma_n) U^T \tag{9}$$

Matrix U is called the Eigenvector matrix of P and $\sigma_1, \dots, \sigma_n$ which are positive definite Eigenvalues of P. Now, the exponential matrix can be formulated as:

$$\exp(P) = U diag(\exp(\sigma_1), \dots, \exp(\sigma_n)) U^T \tag{10}$$

Additionally, the logarithmic matrix can be written as below:

$$\log(P) = U diag(\log(\sigma_1), \dots, \log(\sigma_n)) U^T \tag{11}$$

The space over P is a differentiable Riemannian manifold $\mathcal{M}$ with the dimension of $m = n(n+1)/2$. Derivation of Matrix P over this manifold is a vector space which is called tangent space $T_P$ at a specified point.

Each Riemannian manifold has its Riemannian measure, which is defined as the inner product on every tangent space:

$$\langle S_1, S_2 \rangle_P = Tr(S_1 P^{-1} S_2 P^{-1}) \tag{12}$$

Equation (12) induces the norm for tangent vectors in the form of: $\langle S, S \rangle_P = \|S\|_P^2 = Tr(SP^{-1}SP^{-1})$. In the unitary matrix, this norm is simplified to Frobenius norm $\langle S, S \rangle_I = \|S\|_F^2$.

### 3.1.1. Riemannian distance

Assume $\Gamma(t): [0,1] \to P(n)$ is every curve on the manifold, which joins to different points on the same manifold. The length curve is calculated as:

$$L\left(\Gamma(t)\right) = \int_0^1 \left\|\Gamma'(t)\right\|_{\Gamma(t)} dt \tag{13}$$

The shortest curve on the manifold which connects to points $P_1$ and $P_2$ is called Geodesic. Hence, the geodesic length is Riemannian distance and is calculated as follows:

$$\delta_R(P_1, P_2) = \|\log(P_1^{-1} P_2)\|_F = [\Sigma_{i=1}^{n} log^2 \lambda_i]^{1/2} \tag{14}$$

The $\|\cdot\|_F$ is Frobenius norm and $\lambda_i$ are positive definite Eigenvalues of the matrix $P_1^{-1} P_2$. Some important properties of Riemannian distance are:

- $\delta_R(P_1, P_2) = \delta_R(P_2, P_1)$
- $\delta_R(P_1, P_2) = \delta_R(P_1^{-1}, P_2^{-1})$
- $\delta_R(P_1, P_2) = \delta_R(W^T P_1 W, W^T P_2 W), \forall W \in Gl(n)$

### 3.1.2. Exponential and logarithmic mapping

For each point P in a manifold, there is a tangent space including tangent vectors at point P. Every tangent vector $S_i \in S(n)$ is the geodesic derivation $\Gamma_i(t)$ in $t = 0$ which connects $P$ and $P_i$. Thus, we have:

$$Exp_P(S_i) = P_i = P^{1/2} \exp(P^{-1/2} S_i P^{-1/2}) P^{1/2} \tag{15}$$

$$Log_P(P_i) = S_i = P^{1/2} \log(P^{-1/2} P_i P^{-1/2}) P^{1/2} \tag{16}$$

Equations (15) and (16) are depicted visually in the following figure. It is noticeable that $S_i$ is a tangent vector and belongs to $S(n)$. The $S_i$ is converted to a vector with the dimension of $m = n(n+1)/2$. Moreover, in the symmetric matrix $S_i$, there are only $m$ different values.

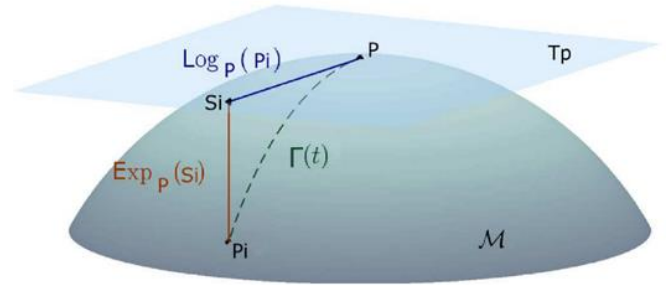

**Fig. 3.** Manifold M and tangent space T_P in point P with exponential and logarithmic operators

There is also a redefinition for Riemannian distance, which comes from the logarithmic mapping concept:

$$\delta_R(P, P_i) = \|Log_P(P_i)\|_P = \|S_i\|_P = \left\|upper(P^{-\frac{1}{2}}\log(P_i)\,P^{-1/2})\right\|_2 \tag{17}$$

The term of $\|\cdot\|_P$ shows the norm of a tangent vector $S_i \in S(n)$ and is calculated as $\|S_i\|_P = \sqrt{Tr(S_i P^{-1} S_2 P^{-1})}$ and additionally, the $\|\cdot\|_2$ is $L_2$ norm of any vector. The operator $upper$ is applied on a matrix to extract the corresponding vector in a way with multiplying diagonal elements which are weighted 1 and also multiplying non-diagonal elements with the weight of $\sqrt{2}$. Thus, $s_i$ is the m-dimensional normalized vector in tangent space, which formulated as:

$$s_i = upper(P^{-1/2}\log(P_i)\,P^{-1/2}) \tag{18}$$

Using equation (16) and (18), we have:

$$s_i = upper(P^{-1/2}S_iP^{-1/2}) = upper(P^{-1/2}P^{1/2}\log(P^{-1/2}P_iP^{-1/2})P^{1/2}P^{-1/2}) \tag{19}$$

Due to $P^{-1/2}P^{1/2} = P^{1/2}P^{-1/2} = I(n)$ and $I(n)$ is a unitary matrix; we have:

$$s_i = upper(\log(P^{-1/2}P_iP^{-1/2})) \tag{20}$$

### 3.1.3. Riemannian mean

Some Riemannian tools need means over symmetric positive definite matrixes. Via definition of Euclidian distance on $\mathcal{M}(n)$, $\delta_E(P_1, P_2) = \|P_1 - P_2\|_F$, the Euclidian mean of a symmetric positive definite matrix is calculated as:

$$\mathcal{A}(P_1, \dots, P_I) = \operatorname{argmin}_{P \in P(n)} \sum_{i=1}^{I} \delta_E^2(P, P_i) = \frac{1}{I}\sum_{i=1}^{I} P_i \tag{21}$$

Riemannian mean is a geometric mean of a set of symmetric positive definite matrixes in $\mathcal{M}(n)$. With Riemannian distance definition, we can formulate Riemannian mean for symmetric positive definite matrixes as below:

$$\mathcal{G}(P_1, \dots, P_I) = \operatorname{argmin}_{P \in P(n)} \sum_{i=1}^{I} \delta_R^2(P, P_i) \tag{22}$$

It is guaranteed that there is a local minimum for the above equation because $P(n)$ is non-positive sectional curvature. Of course, there is no closed form to represent Riemannian mean, so it has to be calculated by iterative optimization algorithms. For example, the following algorithm shows Riemannian mean calculation iteratively:

---

**Algorithm1:** Geometric mean of matrixes $P_1, \dots, P_K$

---

**Input:** K Matrix $P_1, \dots, P_K$
**Output:** Riemannian mean of $P_i$ which $M$
- Initialization of $M$ with random values or with an arithmetic mean
- While $(\|\sum_k \ln(P_k^{-1}M)\|_F > \varepsilon)$:
$$M \leftarrow M^{\frac{1}{2}}\exp\left[\frac{1}{K}\sum_k \ln\left(M^{-\frac{1}{2}}P_iM^{-\frac{1}{2}}\right)\right]M^{\frac{1}{2}}$$
End

---

## 3.2. Classification algorithm based on Riemannian geometry

### 3.2.1. Minimum distance to Riemannian mean

This type is the simplest form of Riemannian classifier, which shortly called MDRM [9]. The algorithm works in a supervisory way. This classifier in training phase calculates within a class means using training dataset then, in test phase calculates Riemannian distance between each test sample and each calculated within class means. Hence, each sample belongs to the group with minimum distance.

### 3.2.2. LDA in tangent space

Although MRDM has earned acceptable results in different tasks [9], to improve performances, researchers decided to use Riemannian geometry in new classifiers that are more complicated. Due to most common algorithms (such as SVM and LDA) works based on mapping to hyperplanes, so directly using symmetric covariance matrix (SCM) properties on Riemannian manifolds is not possible. In this regard, tangent space LDA (TSLDA) is another method, which tries to work with SCM properties.

The main idea in TSLDA is to map SCM samples to Riemannian mean calculated on the training set. Since tangent space is a Euclidian space, it is possible to use LDA technique. To label any input sample, firstly it is mapped into tangent space over the mean of samples, and then the result is vectorized and finally is assigned in accordance with the following equation:

$$y_i = sign(w^T s_i + b) \tag{23}$$

In this equation, $w$ and $b$ are weights and bias respectively which are found by LDA and $s_i$ is mapped vector over the mean value of training samples.

## 4. Results

In this paper, our dataset includes 260 color images with the size of $576 \times 720$ which were acquired from the Research Center of Oncology, Hematology, and Bone Marrow Transplantation of Imam Khomeini in Tehran. For the aim of WBCs recognition, we have considered five different classes, which represent all WBCs, including neutrophils, eosinophil, basophils, lymphocytes, and monocytes. In the following tables, classification accuracy values for different WBCs classes are shown. As depicted in Table, covariance matrix features and TSLDA ensemble can remarkably categorize different WBCs.

| | Basophil | Eosinophil | Lymphocyte | Monocytes | Neutrophils | Accuracy |
|---|---|---|---|---|---|---|
| Basophil | 45 | 4 | 0 | 5 | 1 | **81%** |
| Eosinophil | 5 | 24 | 1 | 7 | 2 | **61%** |
| Lymphocyte | 1 | 2 | 56 | 2 | 0 | **91%** |
| Monocytes | 13 | 9 | 2 | 22 | 2 | **45%** |
| Neutrophils | 0 | 2 | 0 | 1 | 54 | **94%** |

**Table 1.** Confusion matrix with D-SIFT features

| | Basophil | Eosinophil | Lymphocyte | Monocytes | Neutrophils | Accuracy |
|---|---|---|---|---|---|---|
| Basophil | 53 | 0 | 2 | 0 | 0 | **96.36%** |
| Eosinophil | 0 | 35 | 2 | 2 | 0 | **89.74%** |
| Lymphocyte | 0 | 1 | 58 | 1 | 0 | **95.08%** |
| Monocytes | 0 | 2 | 3 | 42 | 0 | **89.36%** |
| Neutrophils | 0 | 2 | 0 | 2 | 53 | **92.98%** |

**Table 2.** Confusion matrix with PHOW features

| | Basophil | Eosinophil | Lymphocyte | Monocytes | Neutrophils | Accuracy |
|---|---|---|---|---|---|---|
| Basophil | 55 | 0 | 0 | 0 | 0 | **100%** |
| Eosinophil | 0 | 39 | 0 | 0 | 0 | **100%** |
| Lymphocyte | 0 | 0 | 60 | 1 | 0 | **98.36%** |
| Monocytes | 0 | 0 | 3 | 44 | 0 | **93.62%** |
| Neutrophils | 0 | 0 | 0 | 0 | 57 | **100%** |

**Table 3.** Confusion matrix with covariance matrix features

| | Segmentation type | Classification | # Images | Accuracy |
|---|---|---|---|---|
| **Our proposed method** | Naïve Bayes | Tangent space LDA | 260 | **98.02%** |
| **Rezatofighi et al. [30].** | Gram-Schmidt orthogonalization and snake | SVM | 400 | **86.10%** |
| **Young [5]** | Histogram threshold | Distance classifier | 199 | **92.46%** |
| **Bikhet et al. [28]** | Entropy threshold and iterative threshold | Distance classifier | 71 | **90.14%** |
| **Tabrizi et al. [29]** | Gram-Schmidt orthogonalization and snake | LVQ | 400 | **94.10%** |

**Table 4.** Comparison among different methods in WBCs recognition task

## 5. Conclusion

In the current paper, we introduced a reasonable procedure, which can recognize WBCs. Recognition power and accuracy of our method were evaluated over 260 different color images, which indicates our method of excellence in comparison to other methods.